\documentclass{vgtc}                          
\vgtcinsertpkg
\ifpdf
  \pdfoutput=1\relax                   
  \usepackage{graphicx}                
  \DeclareGraphicsExtensions{.pdf,.png,.jpg,.jpeg} 
\else
  \usepackage{graphicx}                
  \DeclareGraphicsExtensions{.eps}     
\fi%

\graphicspath{{figures/}{pictures/}{images/}{./}} 

\usepackage{microtype}                 
\PassOptionsToPackage{warn}{textcomp}  
\usepackage{textcomp}                  
\usepackage{mathptmx}                  
\usepackage{times}                     
\usepackage{cite}                      
\usepackage{tabu}                      
\usepackage{booktabs}                  

\usepackage[verbose]{wrapfig}
\usepackage{fontawesome}
\usepackage{adjustbox}
\usepackage[shortlabels, inline]{enumitem}
\usepackage[autostyle=true]{csquotes}
\usepackage{subcaption}
\usepackage{calc}
\usepackage{amsmath}

\usepackage[usenames,dvipsnames]{xcolor}
\usepackage{todonotes}

\definecolor{light-gray}{gray}{0.95}
\definecolor{vianagreen}{rgb}{0.4,0.624,0.4}
\definecolor{vianared}{rgb}{1,0.4,0.4}
\definecolor{vianayellow}{rgb}{1,0.839,0.4}

\definecolor{confirmedblue}{HTML}{47acdc}
\definecolor{suggestionblue}{HTML}{7FDBFF}
\definecolor{suggestiongreen}{HTML}{39CCCC}

\newlength\myheight
\newlength\mydepth
\settototalheight\myheight{Xygp}
\newcommand*\inlinegraphics[1]{%
  \settototalheight\myheight{Xygp}%
  \settodepth\mydepth{Xygp}%
  \raisebox{-\mydepth}{\includegraphics[height=\myheight]{#1}}%
}

\usepackage{lipsum}

\onlineid{0}
\vgtccategory{Research}
\vgtcpapertype{application/design study}

\title{VIANA: Visual Interactive Annotation of Argumentation}
\author{Fabian Sperrle\thanks{e-mail: {firstname}.{lastname}@uni-konstanz.de} %
\and Rita Sevastjanova$^*$%
\and Rebecca Kehlbeck$^*$ %
\and Mennatallah El-Assady$^*$
}
\affiliation{\scriptsize University of Konstanz}

\shortauthortitle{Sperrle \MakeLowercase{\textit{et al.}}: \emph{VIANA}: Visual Interactive Annotation of Argumentation}

\abstract{Argumentation Mining addresses the challenging tasks of identifying boundaries of argumentative text fragments and extracting their relationships.  Fully automated solutions do not reach satisfactory accuracy due to their insufficient incorporation of semantics and domain knowledge. Therefore, experts currently rely on time-consuming manual annotations.
In this paper, we present a visual analytics system that augments the manual annotation process by automatically suggesting which text fragments to annotate next. The accuracy of those suggestions is improved over time by incorporating linguistic knowledge and language modeling to learn a measure of argument similarity from user interactions. 
Based on a long-term collaboration with domain experts, we identify and model five high-level analysis tasks. We enable close reading and note-taking, annotation of arguments, argument reconstruction, extraction of argument relations, and exploration of argument graphs. To avoid context switches, we transition between all views through seamless morphing, visually anchoring all text- and graph-based layers.
We evaluate our system with a two-stage expert user study based on a corpus of presidential debates. The results show that experts prefer our system over existing solutions due to the speedup provided by the automatic suggestions and the tight integration between text and graph views.}


\teaser{
    \centering
    \includegraphics[width=\textwidth]{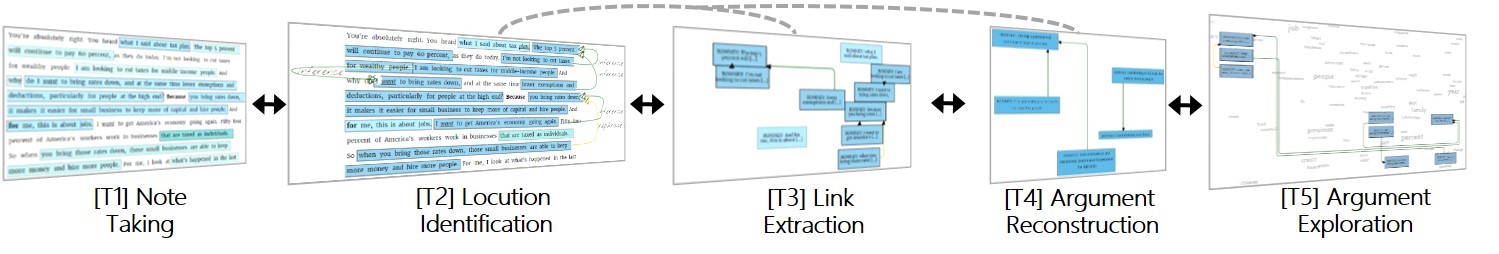}
    \caption{\emph{VIANA} is a system for interactive annotation of argumentation. It offers five different analysis layers, each represented by a different view and tailored to a specific task. The layers are connected with semantic transitions. With increasing progress of the analysis, users can abstract away from the text representation and seamlessly transition towards distant reading interfaces.}
    \label{fig:task_layers}
}

\global\hyphenpenalty=0
\global\linepenalty=5000
\global\looseness=-1
\global\tolerance=10

\begin{document}
\setlength{\intextsep}{0pt}%

\firstsection{Introduction}

\maketitle
Argument mining is a flourishing research area that enables various novel, linguistically-informed applications like semantic search engines, chatbots or human-like discussion systems, as convincingly demonstrated by IBM's project debater~\cite{Slonim2018ProjectDebater}. To achieve reliable performance in these complex tasks, modern systems rely on the analysis of the underlying linguistic structures that characterize successful argumentation, rhetoric, and persuasion. Consequently, to distill the building blocks of argumentation from a text corpus, it is not sufficient to employ off-the-shelf Natural Language Processing techniques~\cite{Wachsmuth2017ArgumentationPractice}, which are typically developed for coarser analytical tasks (see~\cite{liu2018bridging} for an overview), such as with the high-level tasks of topic modeling~\cite{El-Assady2019VisualExecution} or sentiment analysis~\cite{bembenik_towards_2016}.

\begin{figure}[t]
    \centering
    \includegraphics[width=\linewidth]{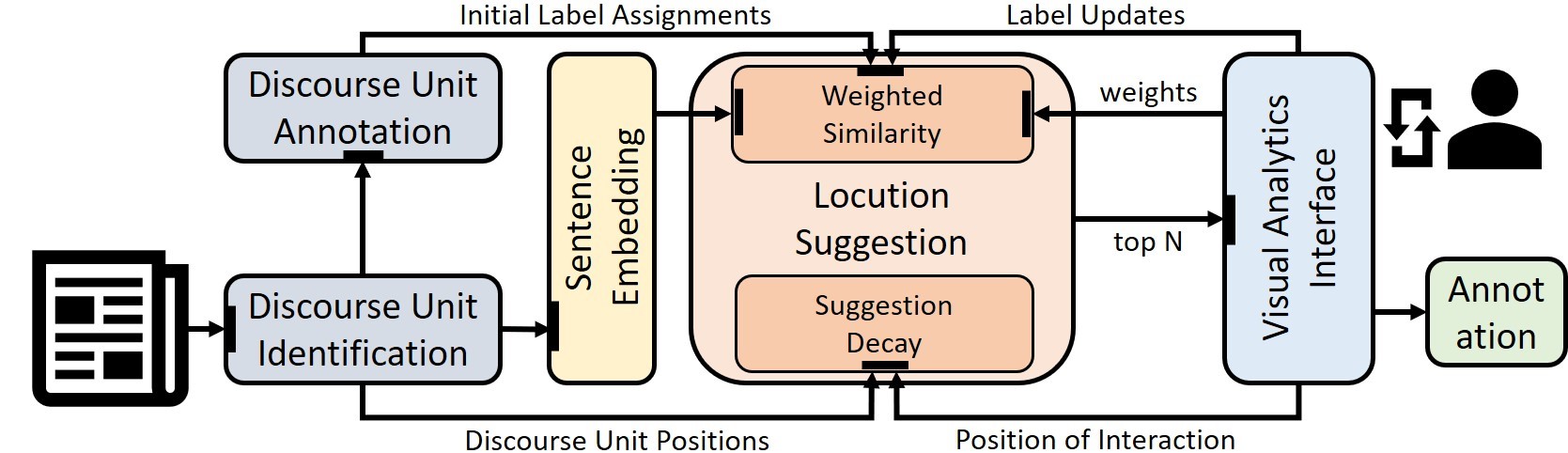}
    \caption{The locution suggestion workflow combines linguistic annotations with a language model and learns from presence and absence of user interactions to improve suggestion quality over time. }
    \label{fig:suggestion_workflow}
    \vspace{-\baselineskip}
\end{figure}

Hence, to master the challenge of identifying argumentative substructures in large text corpora, computational linguistic researchers are actively developing techniques for the extraction of argumentative fragments of text and the relations between them~\cite{Lippi2016ArgumentationTrends}. 
To develop and train these complex, tailored systems, experts rely on large corpora of annotated gold-standard training data. However, these training corpora are difficult and expensive to produce as they extensively rely on the fine-grained manual annotation of argumentative structures. 
An additional barrier to unifying and streamlining this annotation process and, in turn, the generation of gold-standard corpora is the subjectivity of the task.
A reported agreement with Cohen's $\kappa$~\cite{Cohen1960AScales} of~$0.610$\cite{Visser2018RevisitingComparison} between human annotators is  considered ``substantial''~\cite{Landis1977TheData} and is the state-of-the-art in the field.
However, for the development of automated techniques, we have to rely on the extraction of decisive features. 
Cabrio et al.~\cite{Cabrio2013FromDifferences} present a mapping between \textit{discourse indicators} and \textit{argumentation schmes}, indicating a promising direction for more automation.
We use such automatically extracted discourse indicators as a reliable foundation for annotation-guidance. The resulting visual analytics workflow is presented in \autoref{fig:suggestion_workflow}. After discourse units have been annotated with discourse indicators and enriched with sentence-embedding vectors, they are used to train a measure of argument similarity. This measure is updated over time as users annotate more text. To speed up training and remove clutter from the visual interface we introduce a \emph{novel, viewport-dependent approach to suggestion decay}.

Including machine learning and visual analytics into annotation-systems presents a substantial step towards semi-automated argumentation annotation. Systems can rely on a bi-directional learning loop to improve performance: first, they can learn from the available input data, providing both better recommendations and guidance for users. Second, systems can also learn from user interactions to improve and guide the used machine learning algorithms.
Tackling these challenges and employing such progressive mixed-initiative learning, we present a novel visual analytics approach for argumentation annotation in this paper. We base our design choices on a long-term collaboration with experts from the humanities and social sciences. We observed their work processes and underlying theories to gain insight into their respective fields~\cite{Hinrichs2017RiskVisualization}. The well-established Inference Anchoring Theory~\cite{Budzynska2011WhenceInference} (IAT) provides the solid foundation of a theoretical framework that is capable of representing argumentative processes. Having acquired direct insight into argumentation from our collaborations, we present a requirement and task analysis that informs the development of \emph{VIANA}, our annotation system, as well as future approaches in \autoref{sec:req_analysis}.

\textbf{Contributions -- } While we present \emph{VIANA} in the context of the \textit{Inference Anchoring Theory} in this paper, its concepts are readily adaptable to other domain-specific theories and annotation problems, for example from linguistics or political sciences. Thus, this paper's contribution is two-fold.
(i)~We contribute a \textbf{requirement- and task-analysis} for effectively developing visual analytics systems in the field of argumentation annotation. (ii)~We further contribute the \textbf{visual analytics application}, \emph{VIANA}, including a \textbf{novel design of layered visual abstractions} for a targeted analysis through semantic transitions, as well as a \textbf{recommendation system} learning from both domain knowledge and user interaction, introducing \textbf{viewport-dependent suggestion decay}.

\section{Related Work}
Recent years have seen a rise of interactive machine learning~\cite{Fails2003InteractiveLearning} and such techniques are now commonly integrated into visual analytics systems, as recently surveyed by Endert et al.~\cite{Endert2017TheAnalytics}.
Often, they are used to learn model refinements from user interaction~\cite{El-Assady2018} or provide \emph{semantic interactions}~\cite{Endert2012SemanticAnalytics}. Semantic interactions are typically performed with the intent of refining or steering a machine-learning model. In \emph{VIANA}, expert users perform \emph{implicit} semantic interactions, as their primary goal is the annotation of argumentation. The result is a concealed machine teaching process~\cite{Simard2017MachineSystems} that is not an end in itself, but a ``by-product'' of the annotation. 

\textbf{Close Reading and Annotation Interfaces -- }
In their survey, J{\"a}nicke et al.~\cite{Janicke2015OnChallenges} present an overview of visualization techniques which support close and distant reading tasks. According to the authors, ``close reading retains the ability to read the source text without dissolving its structure.''~\cite{Janicke2015OnChallenges} Distant reading generalizes or abstracts the text by presenting it using global features.

Several systems combine the close and distant reading metaphors to provide deeper insights into textual data, such as \emph{LeadLine}~\cite{Dou2012b} or \emph{EMDialog}~\cite{Hinrichs2008}.
Koch et al.~\cite{Koch2014VarifocalReaderDocuments} have developed a tool called \emph{VarifocalReader}, which combines focus- and context-techniques to support the analysis of large text documents. The tool enables exploration of text through novel navigation methods and allows the extraction of entities and other concepts. \emph{VarifocalReader} places all close and distant-reading views next to each other, following the \emph{SmoothScroll} metaphor by W\"orner and Ertl~\cite{Worner2013SmoothScroll:Slider}. \emph{VIANA} instead ``stacks'' the different views into task-dependent layers.

In recent years, several web-based interfaces have been created to support users in various text annotation tasks. For example, \emph{BRAT}~\cite{Stenetorp2012BRAT:Annotation} can be used for the annotation of POS tags or named entities. In this interface, annotations are made directly in the text by dragging the mouse over multiple words or clicking on a single word. \emph{VIANA} employs the same interactions for text annotation. 
Another web-based annotation tool is called \emph{Anafora}~\cite{Chen2013}; it allows annotations of named entities and their relations. Lu et al.~\cite{lu2017visual} use automatic entity extraction for annotating relationships between media streams. TimeLineCurator~\cite{fulda2015timelinecurator} automatically extracts temporal events from unstructured text data and enables users to curate them in a visual, annotated timeline. Bontcheva et al.~\cite{Bontcheva2013} have presented a collaborative text annotation framework and emphasize the importance of pre-annotation to significantly reduce annotation costs. Skeppstedt et al. have presented a framework that creates BRAT-compatible pre-annotations~\cite{Skeppstedt2016PALLearning} and discuss (dis-)advantages of pre-annotation. 
The initial suggestions of \emph{VIANA} could be seen as pre-annotations, but are automatically updated after each interaction.

\textbf{Argument Annotation -- }
Scheuer et al.~\cite{Scheuer2010Computer-supportedArt} offer a comprehensive overview of computer-supported argumentation systems. They characterize five visual argument representations, including graph views, and focus on both systems that allow students to practice the rules of argumentation and those that incorporate collaboration.

\emph{Araucaria}~\cite{Reed2004Araucaria:Representation.} and its more recent online variant OVA+~\cite{Janier2014OVA+:Interface} support the interactive diagramming of argument structures. OVA+, the de-facto standard for argumentation annotation, offers a text view and a graph view side by side. It supports a vast set of argumentation theories and their peculiarities. When annotating, users create detailed argument graphs (see \autoref{fig:ova_example}) through text selection and align them with drag and drop, or rely on a rudimentary automatic layout engine. \emph{Monkeypuzzle}~\cite{Douglas2017MonkeypuzzleTools} relies on the user interface and interactions introduced in \emph{Araucaria}, but adds the possibility to simultaneously annotate texts from multiple sources. Like \emph{VIANA}, both OVA and Araucaria enable annotation according to IAT. However, \emph{VIANA} automatically aligns extracted argumentation graphs with the text view and provides automatically updating suggestions to speed up the annotation process.
Stab et al.~\cite{Stab2014ArgumentationPerspective} have created a web-based annotation tool, combining a text and graph view side by side. Users create arguments directly in the transcript, and each component is assigned an individual color. 
Relations between arguments (attack, support, sequence) are shown in a simple graph structure. 
All introduced systems offer graph and text views and suffer from similar issues. It is usually hard to relate the graph structure to the original text, and large input corpora make the presented information hard to manage. \emph{VIANA} tackles these issues with task-specific interface layers and seamless transitions between text- and graph views. 

\textbf{Interactive Recommender Systems -- } There are generally three approaches to recommender systems: collaborative filtering, content-based, and hybrid approaches~\cite{He2016InteractiveOpportunities}. Collaborative filtering systems utilize ratings or interactions from other users to recommend items~\cite{Herlocker1999AnFiltering,Sarwar2001Item-basedAlgorithms,Linden2003Amazon.comFiltering}, while content-based systems~\cite{Lops2011Content-basedTrends} make predictions purely on attributes of the items under consideration. Hybrid approaches combine both methods. Annotation suggestions by \emph{VIANA} are content-based.  
Various approaches have been developed to react to changing ratings~\cite{Koren2009CollaborativeDynamics, Xiong2010TemporalFactorization} and evolving user preference over time~\cite{Gama2014AAdaptation}. 
Feedback to recommender systems is either explicit, for example in the form of ratings, or implicit, like in the number of times a song has been played or skipped~\cite{Parra2011ImplicitMapping} or how often an item has been preferred over another~\cite{Lerche:2014:UGI:2645710.2645759, Rendle:2009:BBP:1795114.1795167}.
Jannach et al. recently surveyed implicit feedback in recommender systems~\cite{Jannach2018}.
\emph{VIANA} accepts both explicit (acceptance or rejection of a suggestion) and implicit feedback. To incorporate implicit feedback we propose a novel approach to recommendation decay. Based on the current viewport \emph{VIANA} penalizes those suggestions that are visible, but ignored. Penalties increase with decreasing pixel distance between the suggestion and a user interaction. Previous work in recommender systems has often focussed on temporal influence decay~\cite{Isik2017APlatforms,Liu2012Time-BasedFiltering} to lower to influence of older actions on current recommendations. 

\section{Background: Argumentation Annotation}
\label{sec:arg_annot}
\begin{figure}[b]
    \centering
    \vspace{-\baselineskip}
    \includegraphics[width=\columnwidth]{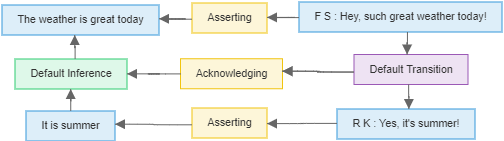}
    \caption{Simple example of an IAT (Inference Anchoring Theory~\cite{Budzynska2011WhenceInference}) structure as produced by OVA+~\cite{Janier2014OVA+:Interface}.}
    \label{fig:ova_example}
\end{figure}
The study of argumentation in political discourse has a history that spans over 2000 years. Since the foundational theories of Aristotle~\cite{Aristoteles1991OnDiscourse}, scholars have been studying the building blocks of successful argumentation and methods of persuasion. This research evolved from a mostly theoretical disputation to the thriving field of data-driven, computational argumentation mining. In a  review of the landmark book ``Argumentation Machines: New Frontiers in Argumentation and Computation''~\cite{Reed2003ArgumentationComputation}, Zukerman defines argumentation as the \enquote{\emph{study of different aspects of (human) interactions whose objective is to reach a conclusion about the truth of a proposition or the adoption of a course of action.}}~\cite{Zukerman2005BookComputation} 
To introduce the specific terminology of the field, we provide a simplified example usage of our system. The expert user's goal is the extraction of argumentative structures and their relations; data that is typically presented in Inference Anchoring Theory (IAT) graphs, as shown in \autoref{fig:ova_example}. To gather the underlying data, she begins by close reading the text and annotating fragments (called \emph{locutions}) of text. In the example, the text is a short discussion between \emph{FS} and \emph{RK} about the weather. The extracted locutions form the right-hand side of the graph and are connected with a \emph{transition}, showing their logical connection. \emph{VIANA} creates those transitions automatically based on the temporal order of locutions. Typically, locutions have exactly one associated \emph{proposition} (left-hand side of the graph) that is also automatically created by the system. Propositions include, for example, premises and conclusions and form the building blocks of arguments.
Having identified locutions, the expert identifies a support relation--called \emph{inference}--between the associated propositions. Alternative types of interpropositional relations are attacks, also called \emph{conflict}, and \emph{rephrases}.
Next, the expert user \emph{reconstructs} the propositions. Reconstruction entails, for example, correction of grammatical issues caused by the extraction of text fragments from their context or capitalization. Consequently, the propositional content differs slightly from the locution. The yellow boxes in \autoref{fig:ova_example} are called \emph{illocutionary connectors} and form a cornerstone of the IAT. While the connections are automatically created with their associated left-hand sides of the graph, the user proceeds to select the labels from a wide array, including ``asserting'', ``questioning'' or ``challenging''. Having completed the annotation she proceeds to an overview map showing the different concepts and topics contained in the propositions to validate her results.

Apart from the simplistic example presented above, \emph{VIANA} can be used to annotate more complex IAT scenarios that can only be introduced very briefly here due to space limitations; 
the supplementary material provides more detailed explanations.
\emph{Linked Arguments} describe propositions that can only create an inter-propositional relation together and not on their own. Instead of attacking a proposition, \emph{undercuts} attack inter-propositional relations. \emph{Indexicals}~\cite{Budzynska2014} are locutions like ``Of course not!'' that rely on the content of the previous locution and lose their meaning when separated. Further, IAT resolves reported speech into artificial locutions. 

\textbf{Requirement and Task Analysis--}
\label{sec:req_analysis}
From our long-term collaboration with experts in philosophy and computational linguistics, we identify several requirements for systems tailored to the task of text annotation and, in particular, argumentation annotation. We categorize our collection of requirements into \textit{general} needs and items that are \textit{specific} to the domain of argumentation mining. 

In \textbf{general}, text annotation tools should include \emph{close and distant reading interfaces} to provide ways to work on the text while also allowing to abstract and generate higher-level insights. These interfaces need to be connected in such a way that users can easily switch between them and \emph{avoid unnecessary losses of context}. This includes keeping the interface clean and easy to use, \emph{removing clutter and distractions}. \emph{User guidance} can greatly speed up the analysis process and facilitate the \emph{curation of results} and their exploration. Once users have compiled results or insights, tools should offer ways to \emph{export and share} those results using \emph{visualizations} suitable for communicating the findings to both experts and non-experts.

Systems for an efficient \textbf{argumentation annotation} need to \textbf{[R1]} deal with large amounts of text and \textbf{[R2]} extract graph structures from that text. Experts' requirements for such systems further include the possibility to 
\textbf{[R3]} extract argumentative fragments of text as locutions and 
\textbf{[R4]} reconstruct propositions from them. Once propositions have been extracted, they also need to be able to
\textbf{[R5]} connect propositions with relations like inference and conflict and
\textbf{[R6]} capture argumentation schemes and annotate illocutionary forces.
Furthermore, discussions often revisit previously mentioned topics, necessitating appropriate functionality to 
\textbf{[R7]} connect propositions with large temporal gaps.

From this requirement analysis we derive five abstract, high-level tasks for argumentation annotation, tailored to expert annotators and analysts: 
    \textbf{[T1]} Close Reading and Note-Taking,
    \textbf{[T2]} Text Segmentation and Locution Identification,
    \textbf{[T3]} Relationship Extraction,
    \textbf{[T4]} Argumentation Reconstruction,
    \textbf{[T5]} Argument Exploration.
These tasks need to be supported by systems catering to argumentation annotation. For the design and implementation of \emph{VIANA}, we translate the five tasks to five distinct, interactive views that support completing them: the
\emph{Note Taking}, 
\emph{Locution Identification}, 
\emph{Link Extraction}, 
\emph{Argument Reconstruction}, and 
\emph{Argument Exploration} 
views from \autoref{fig:task_layers} will be introduced in detail in \autoref{sec:design_considerations}.
All views are presented as stacked layers and connected via semantic transitions. With advancing annotation progress, users transition through the layers, increasing the interface abstraction and transitioning from a pure text-based view to a high-level graph abstraction. 

\section{Workspace Design Considerations}
\label{sec:design_considerations}
As introduced in the previous section, our system is grounded in linguistic argumentation theory. While the number of existing annotation systems conforming to the theory is limited, domain experts actively use those available tools. We thus anchored our design decisions in those accepted applications, as they can represent the linguistic theory and have already formed the experts' mental models that are not easily changed now, as our expert user study confirmed. 

From the previous long-term collaboration with said experts, we also gathered several issues with available annotation systems. One frequently mentioned complaint was the lack of a connection between extracted locutions and their context in the original transcript. We overcome this limitation by directly annotating locutions in the transcript by highlighting the respective words. 
To emphasize the ``hand-made'' nature of the annotation, we offer the option to employ sketch-rendering techniques when displaying locution annotations as well as the connections between them to encourage users to keep refining them. While we initially considered mapping the roughness of the sketch to the uncertainty of the annotation we rejected this idea as comparing different levels of ``sketchiness'' is extremely difficult. Wood et al.~\cite{Wood2012} compared ``normal'' and sketchy visualizations for different use cases. They conclude that user engagement increases with sketchy visualizations when compared to non-sketchy ones.  Additionally, they note that the overall interaction with a tool is perceived as more positive if it uses sketchy rendering. Our study did, however, not fully confirm this finding. While some experts appreciated the sketchy design (thanks to the ``hand-made'' look), others rejected it. This feedback prompted us to add a ```sketchiness slider'' after the first phase of the study.
While the application loads without sketchiness by default, users can now select between no~$\vcenter{\hbox{\includegraphics[height=1.3em]{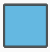}}}$, some~$\vcenter{\hbox{\includegraphics[height=1.3em]{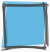}}}$, or strong~$\vcenter{\hbox{\includegraphics[height=1.3em]{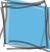}}}$ sketchiness. \autoref{fig:trump_rephrasing} shows the system with sketchiness; all other screenshots include no sketchiness. As sketchiness is employed only for locutions in text-based views the risk of it use visually cluttering the workspace is low, but further research is needed to determine which presentation is most effective~\cite{Behrisch2018Quali-42964}.

The typical size of a corpus annotated for argumentation ranges from 10,000 to 100,000 words. As annotating the transcript of one hour of a debate or discussion can take up to fifty hours, annotators usually split this input into manageable chunks, annotate them separately, and carefully merge the intermediate results. The main reason for chunking the input is that it is otherwise difficult to maintain an overview of what has been annotated. \emph{VIANA} highlights the identified locutions in the input text, enabling experts to relate arguments and their relations to their origin and simplifying the task of keeping an overview. Consequently, annotators can increase the amount of text they load into the application. 

Some existing systems rely on manual node placement on the annotation canvas. With increasing sizes of the argumentation graph, more and more time is spent on keeping the canvas organized. While automated layout routines do exist, they are not always as effective as our experts would expect due to the complexity of IAT graphs. With three different argument graph views using automatic layouts and only showing task-relevant information we free users from this strenuous task, enabling them to focus on annotating instead. 

All text and graph views mentioned above will be introduced in detail in \autoref{sec:interface_layers}. In the following section, we introduce the layering and transitions between these views. 

\subsection{Layered Interface}
\begin{figure}
    \centering
    \adjustbox{cfbox=light-gray 1pt 0pt}{\includegraphics[width=\columnwidth]{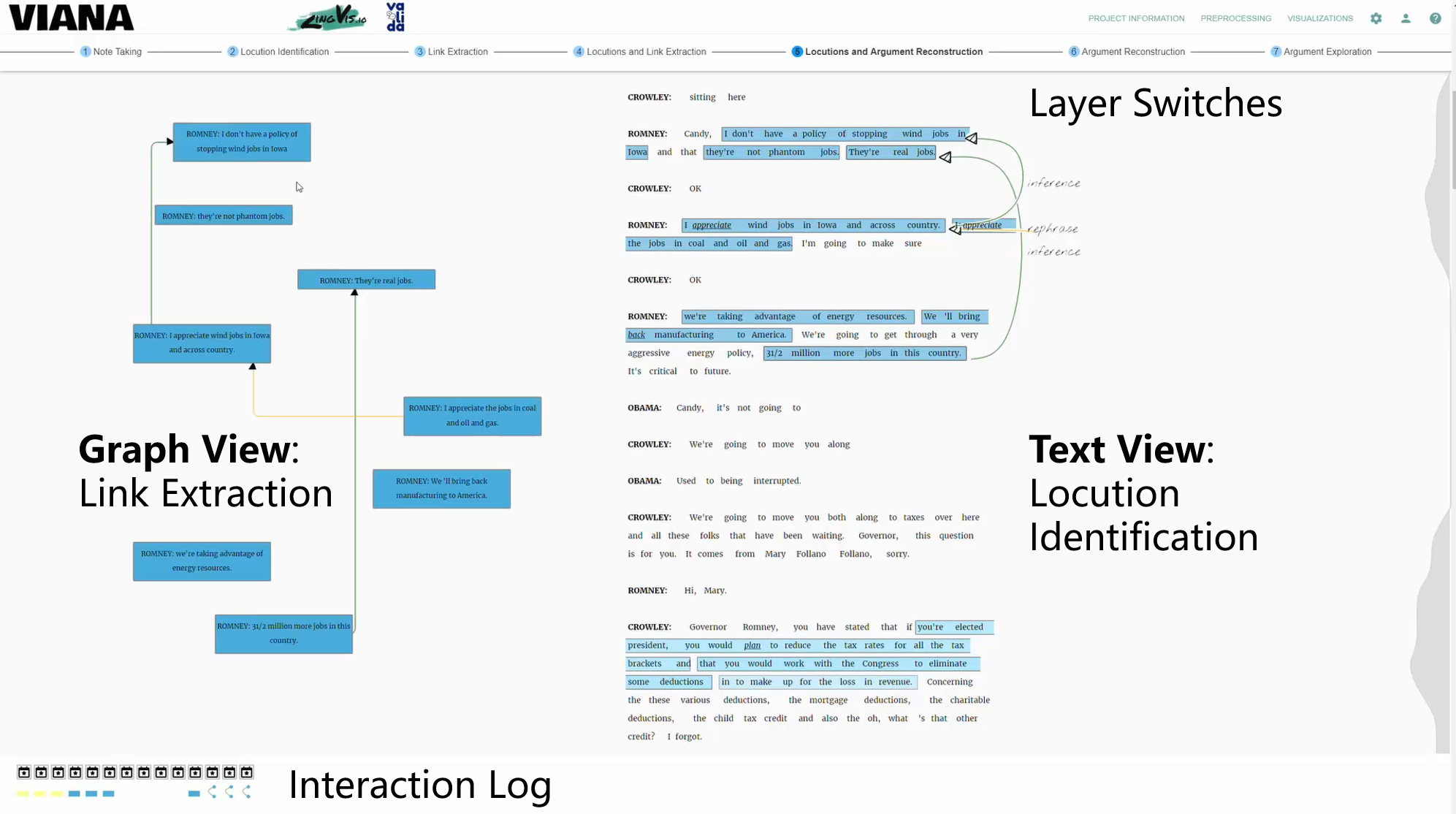}}
    \caption{\emph{VIANA} system overview. The top bar shows the currently active annotation layer. Here, a graph view is shown next to a text view. An interaction log is displayed at the bottom of the screen.}
    \label{fig:system_overview}
    \vspace{-\baselineskip}
\end{figure}

The previous section has already alluded to the typical sizes of argumentation corpora. 
While Scheuer et al. claim that scrolling interfaces can cause users to ``lose the big picture''~\cite{Scheuer2010Computer-supportedArt} they are difficult to avoid in text-based systems. We instead prioritize reducing the amount of information on screen through the introduction of layers. 
By providing the task-specific layers shown in \autoref{fig:task_layers} users only see the information that is currently relevant to them.
By advancing from one task to another on the spectrum, users transition away from text-level views towards a graph-based overview. The different views are intended to enable both close and distant reading. To switch between layers users scroll their mouse wheel while pressing the control key. In order to avoid context switches and make the changes as easy to follow as possible, we smoothly morph all elements on screen. 
The \emph{Note Taking}, \emph{Locution Identification} and \emph{Link Extraction} views are aligned to minimize positional movement. Barring overlap removal, the respective top-left corners of locution annotations and graph nodes are placed at the same screen position, leaving users with morphing, but stationary rectangles and graph edges. When switching to the \emph{Argument Reconstruction} or \emph{Argument Exploration} views, transitions and targeted scrolling support users in keeping the context. To make changes easier to follow, elements under the mouse remain there after the transition, if possible. This concept is familiar from zooming in maps or image viewers. 

As the layers are organized by task progression and often provide functionality for multiple tasks, frequent back-and-forth switches between layers can be avoided. After the first evaluation phase, we added two additional layers at the request of some users. They are indicated by dashed lines in \autoref{fig:task_layers} and contain two visualizations side-by-side to enable parallel work on multiple tasks, at the cost of higher information density. As the individual layers remain available, users can select whichever representation is most effective for them in their current context. The system overview in \autoref{fig:system_overview} presents such a combined layer showing both the link extraction and locution identification views at the same time.

We initially decided to introduce layers rather than employing multiple coordinated views or a tabbed or multi-window interface to facilitate relating the resulting argument graph structure to the original text. Scrolling through the layers maps the graph directly into the original text fragments. While linking and brushing in coordinated views could offer similar functionality, it would require at least three views (propositions, locutions, text). Some expert users in our study preferred the layered approach over multiple parallel views as it enabled them to reduce the amount of information to a level they were comfortable with. 
Heer and Robertson studied animation in statistical data graphics and found that ``animated transitions can significantly improve graphical perception.''~\cite{Heer2007AnimatedGraphics}. We argue that their result ``animation is significantly better than static across all conditions''~\cite{Heer2007AnimatedGraphics} in object tracking tasks is also applicable during layer switches in \emph{VIANA}. Consequently, we employ a layered approach rather than using tabs or multiple windows. 

\subsection{Visual Representation of Illocutionary Connectors}
\begingroup
\setlength{\columnsep}{1em}

Due to the introduction of interface layers, propositions and locutions are not always shown on the screen at the same time. As a result, illocutionary connectors can no longer be rendered like in \autoref{fig:ova_example} (yellow nodes).
They are, however, a fundamental part of the
\begin{wrapfigure}{r}{0.4\linewidth}
\includegraphics[width=\linewidth]{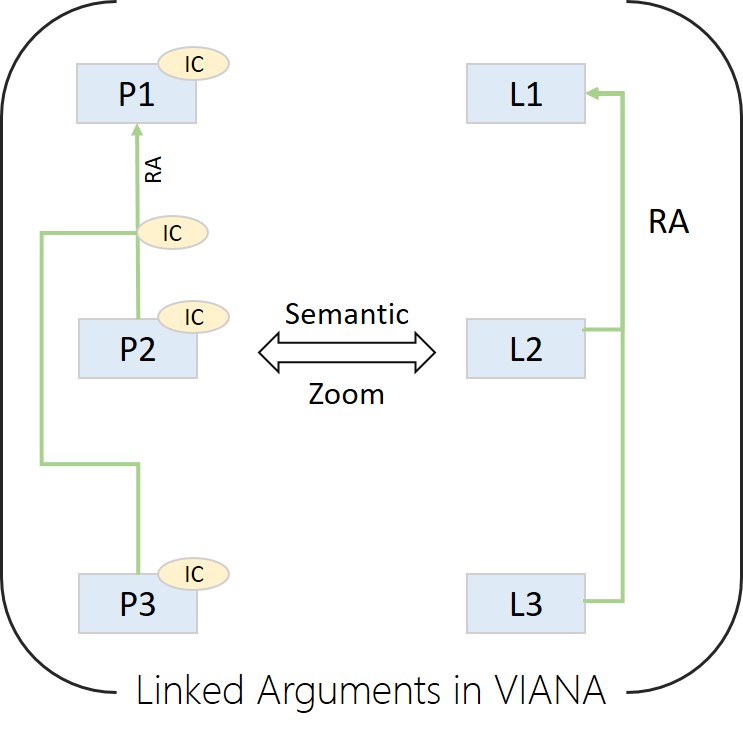}
\end{wrapfigure}
underlying Inference Anchoring Theory and need to be represented. We thus map them to the left-hand side of the graph, showing them as badges on both propositions and interpropositional relations between them.
While the existence of these connectors is fundamental to the theory, the importance of their particular values is task-dependent and they can often be initialized with sensible default values.
\emph{VIANA} thus initializes connections between locutions and propositions as ``Asserting'' and those between transitions and inferences as ``Arguing''.
We provide a setting hiding the illocutionary connectors, allowing users to focus on other tasks and checking the connectors at a different time. 

\phantom{test}\vspace{-\baselineskip}
\endgroup

\begin{figure*}[t]
    \begin{subfigure}[b]{0.4\linewidth}
       \centering
       \adjustbox{cfbox=light-gray 1pt 0pt}{\includegraphics[height=100pt]{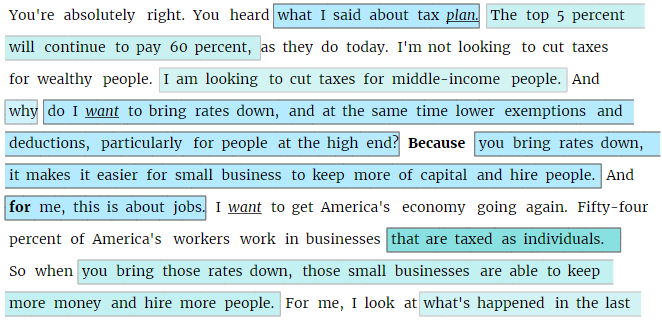}}
       \caption{The \emph{Note Taking View} with proposed annotations}
       \label{fig:slow_analytics_view}
    \end{subfigure}\hfill%
    \begin{subfigure}[b]{0.555\linewidth}
        \centering
        \adjustbox{cfbox=light-gray 1pt 0pt}{\includegraphics[height=100pt]{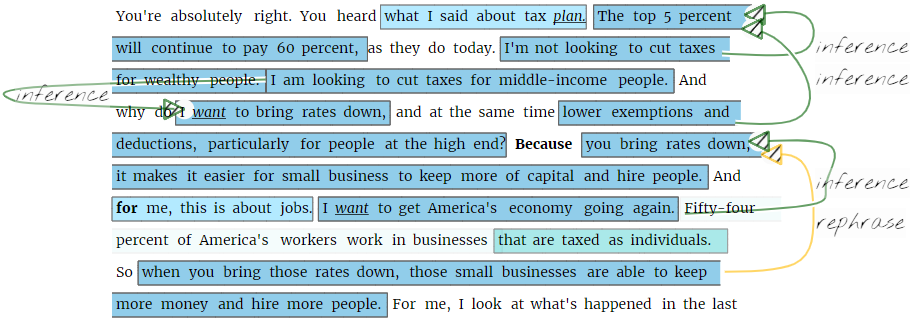}}
        \caption{The \textit{Locution Identification View} showing the result of an annotation}
        \label{fig:text_view}
    \end{subfigure}
    \caption{The two text-based views tailored towards slow analytics and locution identification. In the Slow Analytics View, users can read the text and jot down notes. In the \textit{Locution Identification View}, they can mark locutions and introduce relations between their propositions. }
    \label{fig:figure5}
    \vspace{-\baselineskip}
\end{figure*}
\subsection{Automated Suggestions}
\label{sec:automated_recommendations}
\emph{VIANA} highlights keywords that are of specific interest to the annotation based on pre-defined word lists~\cite{Hautli2017RhetoricalDialogs}.
Connectors like ``so'', ``if'', ``because'' or ``for'' are \textbf{bold} and keywords like ``appreciate'', ``promise'' or ``complain'' that are associated with speech acts are \textit{\underline{italicized and underlined}}. These highlighting-techniques have been found to work comparatively well in the presence of ``distractors'' like the boxes around locutions~\cite{Strobelt2016GuidelinesTechniques}.

In addition to highlighting keywords as guidance for manual annotation, we provide proposed fragments of text that should be annotated as locutions. These fragments are discourse units that have been classified as potential locutions by our recommendation system introduced in \autoref{sec:learning_from_interaction}.
Possible interactions like confirming or rejecting suggested locutions follow guidelines for Human-AI interaction~\cite{Amershi2019GuidelinesInteraction} and will be introduced together with their influence on future suggestions in \autoref{sec:text_view} and \autoref{sec:suggestion_refinement}, respectively. 

\begingroup
\setlength{\columnsep}{1em}

Locutions that have been confirmed by users are shown in a dark blue~$\vcenter{\hbox{\includegraphics[height=1.3em]{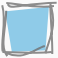}}}$, while those suggested by linguistic rules~$\vcenter{\hbox{\includegraphics[height=1.3em]{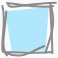}}}$ and predictions based on user-interactions~$\vcenter{\hbox{\includegraphics[height=1.3em]{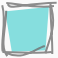}}}$ are light blue and teal, respectively. The certainty of suggested annotations is mapped to their opacity. We deliberately chose shades of blue to avoid conflicts with the colors used for relations between arguments. We selected three relatively similar colors to avoid overwhelming users with too much information; an approach that was validated by experts in our user study. An earlier design of the system colored locutions based on the presence of six different types of discourse unit connectors (see 
\begin{wrapfigure}[3]{r}{0.095\linewidth}
\inlinegraphics{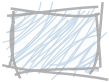}
\inlinegraphics{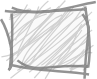}\\
\inlinegraphics{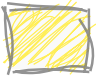}
\inlinegraphics{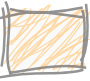} \\
\inlinegraphics{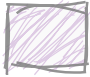}
\inlinegraphics{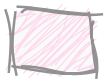}
\end{wrapfigure}
\autoref{sec:initial_suggestions}) 
as shown on the side. 
We chose a simplified color scheme based on expert feedback that such information was interesting but not helpful during the annotation process.

\phantom{blubb}\vspace{-\baselineskip}

\endgroup

\section{Task-Driven Interface Layers}
\label{sec:interface_layers}
In the following section, we introduce the layered views that \emph{VIANA} offers for specific tasks. Several layers offer functionality suited for multiple tasks, and two intermediate layers merge graph- and text views. To transition between layers, users can either select a target layer from a list at the top of their screen or press the control key while using their mouse wheel. 

\subsection{Slow Analytics and Note-Taking}

\begingroup
\setlength{\columnsep}{1em}%

The \emph{Note-Taking and Slow Analytics View} represents the ``distraction free'' mode of \emph{VIANA} and is presented in \autoref{fig:slow_analytics_view}. Depending on user settings, it shows only the raw text or includes the initially proposed locutions. Interviews with experts have revealed different approaches to argument annotation. While one approach starts by immediately annotating locutions, another approach initially scans the text for passages of particular interest. 
The \emph{Slow Analytics View} offers a 
note-taking interface--addressing task [T1] presented in the 
\begin{wrapfigure}{r}{0.25\columnwidth}
\includegraphics[width=\linewidth]{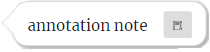}
\end{wrapfigure}
introduction--that allows users to jot down ``free-form'' notes on fragments of text without being forced to mark them as locutions.
This is an advantage over other annotation systems~\cite{Janier2014OVA+:Interface}
where users employ such tactics to compensate for missing functionality.
Once users have gained an overview of the corpus at hand, they progress to the \emph{Text View} for locution identification and relation extraction. 

\phantom{blubb}\vspace{-\baselineskip}

\endgroup

\subsection{Text Segmentation and Locution Identification}
\label{sec:text_view}

Users transition to the \emph{Locution Identification View} to perform [T2] and annotate locutions.
To create a new locution boundary, they select a fragment of the text by clicking and dragging. Once the users let go of the mouse button, both the locution and the corresponding proposition are automatically extracted. The locution is connected to the temporally preceding locution via a transition. However, to avoid cluttering the view, these automatically created transitions are not displayed on the screen and are only contained in the result extracted at the end of the analysis. 
As an alternative to manual annotation, users can explore the proposed locutions. \emph{VIANA} displays them in more muted colors and with a lower opacity, as can be seen in \autoref{fig:slow_analytics_view}. The colors encode the different origins for these fragments as introduced in \autoref{sec:automated_recommendations}. 
An area chart on the right-hand side of the screen summarizes the annotations and can show regions with fewer annotations than expected. 
Those regions might either be of less interest to the analysis or indicate missed locutions.

Every locution displays a toolbar when hovering over it. This toolbar allows confirming~\faicon{check} or unconfirming~\faicon{question} a locution, opening the edit~\faicon{pencil} tooltip to change its type or provide an annotation (as introduced in the slow analytics view), deriving a locution from it~\faicon{comments}, for example, to resolve reported speech, or deleting it~\faicon{times} outright. 

To draw a connection between two locutions, users click and drag from one source locution to a target.
Pressing the shift key while dragging will result in a transition while pressing control will result in an inference. A transition is shown as a light grey line \inlinegraphics{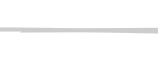} connecting the two locutions directly. 

By default, any non-transition edge is drawn as an inference~\inlinegraphics{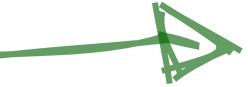}. Double-clicking on the edge or its label iterates through the available edge types, including conflict~\inlinegraphics{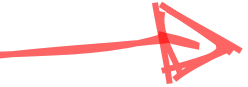} and rephrase~\inlinegraphics{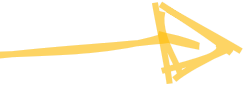}. The associated colors green, red and yellow are well-established in the argumentation community. A toolbar similar to that described for locutions is available for propositional relations as well. It allows to open an edit tooltip \faicon{pencil} or delete the edge \faicon{times}. Users can change the type of edge in the tooltip using a drop-down menu. Users can also set an \emph{argumentation scheme}~\cite{Walton1996ArgumentationReasoning} in the tooltip. Such a scheme describes the type of the relation more precisely and is displayed instead of the default ``inference'', ``conflict'', and ``rephrase'' as an edge-label once selected. Additionally, the tooltip allows users to set the illocutionary connector between the transition and the propositional relation. It will be shown as a badge in the Graph View.

\begingroup
\setlength{\columnsep}{1em}%

\begin{wrapfigure}[4]{r}{0.2\linewidth}
\setlength{\fboxsep}{2pt}%
\setlength{\fboxrule}{0.3pt}%
\adjustbox{cfbox=light-gray 1pt 0pt, center}{\includegraphics[width=\linewidth]{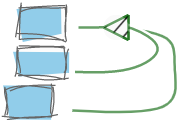}}
\end{wrapfigure}
To create a linked argument, users can draw an edge to the arrowhead of an existing propositional relation. %
This causes the arrowhead to move backward from the end of the edge and receive both incoming edges there. The increased distance between arrowhead and locution emphasizes the merge and clearly distinguishes linked arguments from converging arguments. Converging arguments are achieved by simply drawing multiple edges ending in the same locution. Contrary to linked arguments, the premises of a  converging argument can all support or attack the conclusion individually. Consequently, no unique visual mapping showing their connection is necessary or warranted. 

\begin{wrapfigure}[4]{r}{0.2\linewidth}
\setlength{\fboxsep}{2pt}%
\setlength{\fboxrule}{0.3pt}%
\adjustbox{cfbox=light-gray 1pt 0pt, center}{\includegraphics[width=\linewidth]{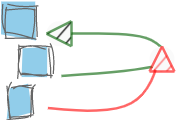}}
\end{wrapfigure}
To create an undercut, i.e., the (typically) attack or support of an existing propositional relation, a new edge is drawn to the label of an existing edge. Users can link up more than two arguments and undercut or support an existing undercutting relation.

\begin{figure*}[t]
    \centering
    \captionsetup[subfigure]{justification=centering}
    \begin{subfigure}[b]{105pt}
        \centering
        \adjustbox{cfbox=light-gray 1pt 0pt}{\includegraphics[height=145pt]{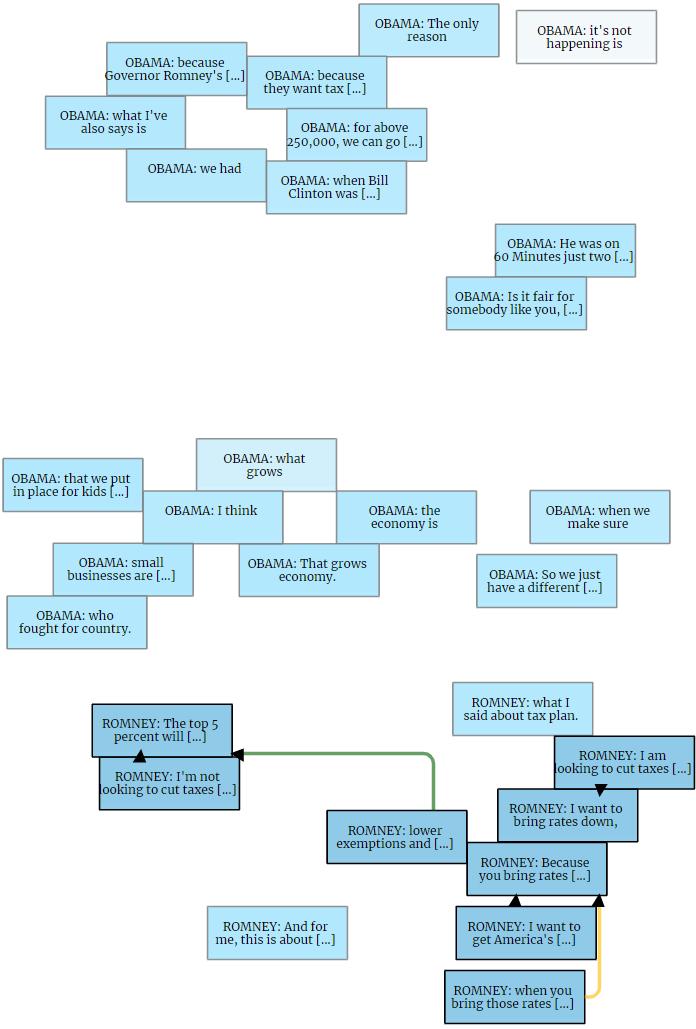}}
        \caption{The \emph{Link Extraction View}\\ with collapsed propositions}
        \label{fig:argument_graph}
    \end{subfigure}\hfill%
    \begin{subfigure}[b]{105pt}
        \centering
        \adjustbox{cfbox=light-gray 1pt 0pt}{\includegraphics[height=145pt]{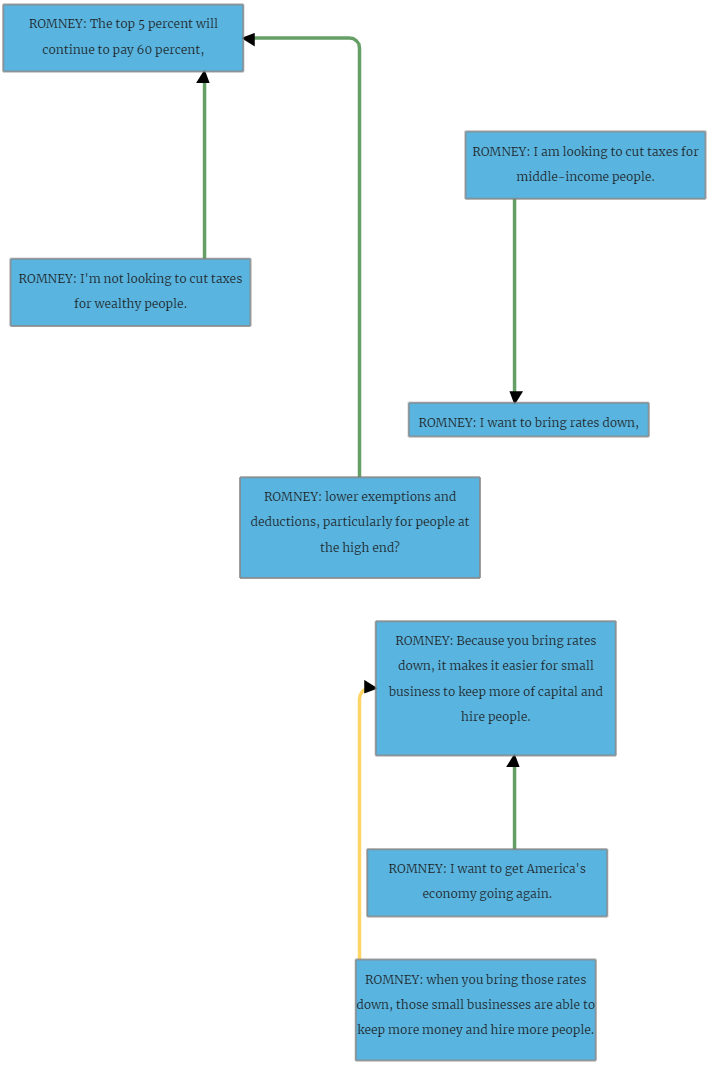}}
        \caption{The \emph{Argument Reconstruction View} with ordered propositions}
        \label{fig:temporal_graph}
    \end{subfigure}\hfill%
    \begin{subfigure}[b]{282pt}
        \centering
        \adjustbox{cfbox=light-gray 1pt 0pt}{\includegraphics[height=145pt]{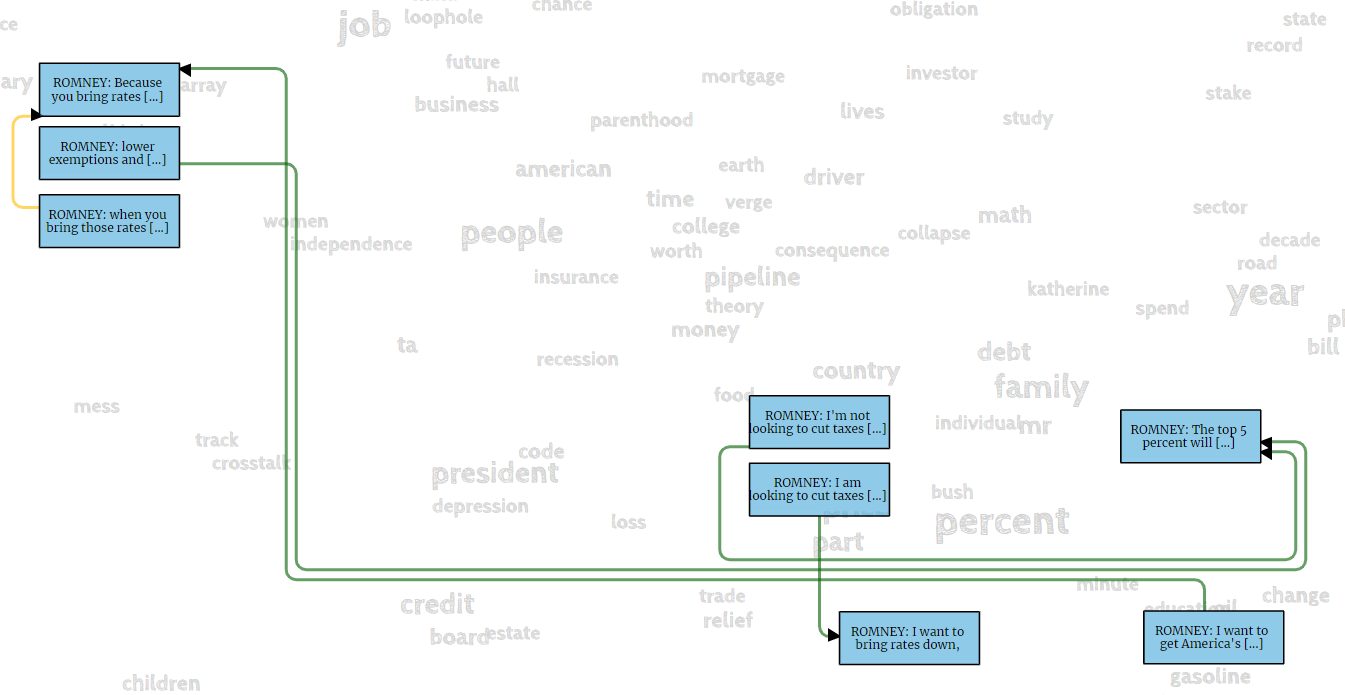}}
        \caption{The \textit{Argument Exploration Map} giving an overview of the result of an annotation. \\This excerpt of the map focuses on topics related to taxes and the economy.}
        \label{fig:topic_map}
    \end{subfigure}
    
    \caption{The graph views enable users to explore the extracted propositions. The \emph{Argument Roncstruction View} temporally orderes propositions (same data as in \autoref{fig:figure5}) and highlights groups and disconnects in the argument graph. The \emph{Argument Exploration Map} provides a distant-reading overview and shows that the annotated text discussed the impact of tax reductions on people.}
    \vspace{-\baselineskip}
\end{figure*}

These simple interactions allow users to annotate the text with locutions without having to switch between separate text and graph visualizations. By staying within the same visualization, the context of the original utterances remains available, facilitating the annotation.
By overplotting the annotations over the text view, the information density on this level rises progressively during the annotation. Nonetheless, we chose this design to prevent users from having to continuously switch layers or visual representations. \autoref{fig:text_view} shows typical density of annotation. \autoref{fig:trump_rephrasing} shows an utterance by Donald Trump and is denser due to his typical style of speech with many repetitions and short arguments rather than long explanations.
Alternative designs would remove either the relation annotation (note taking view) or the text (graph views) to lower the information content displayed on screen and freeing up pixels. As those views are implemented in the system, users are free to choose whichever view fits their workflow best. Such freedom of choice proved important in our user study with experts that have a clear workflow in mind.
\endgroup

\subsection{Relationship Extraction}

While the previously introduced text view also enables the creation of relations, it already contains a lot of information. To enable users to focus on task [T3]--extracting relations between already annotated locutions--\emph{VIANA} contains the \emph{Link Extraction View} shown in \autoref{fig:argument_graph}. The nodes of this graph are propositions (i.e., the left-hand side of an IAT diagram) associated with locutions visible in the text view. To enable smooth transitions between the text and graph view, while avoiding overlap and maintaining readability, this graph shows nodes with shortened text representations at the position of the locution in the text. This positioning allows the outlines of locutions to seamlessly morph into graph nodes, and vice versa. In some cases, minimal position changes are necessary to remove node overlap.

The view provides a quick, compact overview of the current annotations and makes it easy to relate the propositions to their respective locutions as they morph into each other. Hovering over a proposition expands it to fit the entire text. 
Missing propositional relations can be drawn using the same interactions as in the text view.

To prevent overlap issues we utilize a modified version of WebCoLa
by Dwyer et al. based on their previous work on graph layouting~\cite{Dwyer2006IPSep-CoLa:Graphs}. It produces a gridified layout with no node-edge-overlap and fewer edge crossings than drawing direct links between nodes. The remaining edge crossings are easy to make sense of as they are always intersections of orthogonal lines. The same techniques are applied in the two graph views presented in the following sections. 
\subsection{Argument Reconstruction}    
\label{sec:graph_view}

While the \emph{Link Extraction View} still contains the propositions of unconfirmed proposed locutions, they are removed in the \emph{Argument Reconstruction View} shown in \autoref{fig:temporal_graph}. 
All remaining nodes are expanded to fit the entire proposition. To resolve the resulting overlap issues, the nodes are redistributed along the y-axis. Whenever the nodes of two propositions overlap, the one mentioned in the original text at a later time is shifted downwards until the overlap is resolved. The result is an automatically arranged timeline graph of propositions that can be read from the top left to bottom right. This temporal alignment identifies connected components of the graph and facilitates detecting breaks and cuts in the annotation. 

\begingroup
\setlength{\columnsep}{1em}%

Double-clicking a node enables users to edit a proposition in-place. 
Changing the text of a proposition is called \emph{reconstruction} 
\begin{wrapfigure}[6]{r}{0.30\linewidth}
\includegraphics[width=\linewidth]{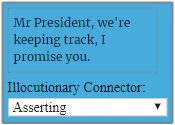}
\end{wrapfigure}
([T4]) and is introduced in \autoref{sec:arg_annot}. The necessary amount of reconstruction varies depending on the analysis task at hand.
Typical changes include rephrasing the locution to form a complete sentence with subject, verb, and objects, lowercasing the first letter, and resolving pronouns. Once users have reconstructed a proposition, the graph nodes show the reconstructed text as their label. The text of the underlying locution is available when hovering over the node. 

In addition to reconstructing propositions, users can also change the illocutionary connector when editing a proposition. This connector is initialized with ``Asserting''. Depending on the proposition other possible values include ``Questioning'', ``Restating'', or combinations like ``Assertive Questioning''. While the temporally ordered graph with expanded propositions is designed for reconstruction in particular, all three graph views available in \emph{VIANA} support the task using the same double-click interaction. Hence, any graph view can also be used to introduce new relations between propositions.
\endgroup

The initial version of \emph{VIANA} showed only one visualization at a given time. After the first study phase we added two layers that combine the \emph{Locution Identification} view on the right and either the \emph{Argument Reconstruction} or \emph{Link Extraction} view on the left, as shown in \autoref{fig:system_overview}. This addition is a direct response to requests from some experts wanting to frequently switch between the tasks.

\subsection{Argument Exploration}
The \emph{Argument Exploration Map} provides a distant reading interface enabling exploration ([T5]) by giving an overview of groups of arguments and their respective important keywords. 
Besides exploration, such a map facilitates the communication of annotation results and enables progress checking during the annotation. It is also useful when continuing the annotation after a short break, as our user study revealed.
The graph shown in this view retains all interactive functionality available in the other graph views.
The possibility to create new relations in the map view is helpful to create so-called ``long-distance relations''. These appear, for example, in political debates where speakers frequently refer back to something that was said some time ago. The topic map reduces the distance between the propositions if they have similar content, facilitating the creation of a relation, and avoiding scrolling. 

To create the map, we first extract all nouns and named entities from the input corpus~\cite{el2019semantic}. Additionally, we gather the top five corpus words each according to word frequence, TF-IDF, $G^2$ and average likelihood ratio. We represent each keyword as a vector using a word embedding tailored to concept words~\cite{Speer2017ConceptNetKnowledge} that combines and optimizes the results of word2vec~\cite{Mikolov2013EfficientSpace} and Glove~\cite{PenningtonJeffrey2014Glove:Representation}.  We then project these high-dimensional word vectors to 2D using tSNE~\cite{vanderMaaten2014AcceleratingAlgorithms} and fit the created space to the dimensions of the screen. As showing all keywords at once would clutter the screen, we only place those on the map that appear in potential suggested locutions (see \autoref{sec:automated_recommendations}).
After scaling keywords according to their frequency in the text, we obtain a map of the most relevant keywords.

We represent each proposition as a weighted average word embedding vector of nouns and named entities. The initial weights for named entities are twice those for nouns to capture their typical importance. If a proposition contains no keywords that are visible on the topic map, we employ a nearest-neighbor search in the word-embedding space to find appropriate, representative nouns. This step is especially crucial for propositions containing no nouns or named entities. Having obtained a representative aggregated embedding vector for a proposition, we can determine its position in the projection. While we prevent node-overlap in the map by nudging nodes apart, we cannot prevent all node-keyword overlap. As a consequence, we offer a toggle switch that swaps the z-order of keywords and nodes to reveal keywords hidden behind nodes.

\subsection{Interaction Log}
\begingroup
\setlength{\columnsep}{1em}%

The interaction timeline at the bottom of the screen summarizes the annotation session and is visible from every layer. 
It shows information about identified \inlinegraphics{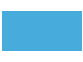} and deleted 
locutions \inlinegraphics{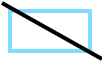}. Employing 
\begin{wrapfigure}[2]{r}{0.50\columnwidth}
    \adjustbox{cfbox=light-gray 1pt 0pt}{\includegraphics[width=\linewidth]{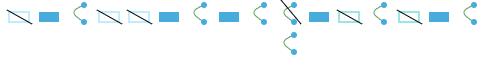}}
\end{wrapfigure}
the same visual metaphors, it also informs about added \inlinegraphics{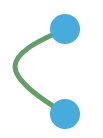}, re-typed \inlinegraphics{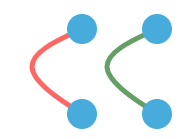} and deleted \inlinegraphics{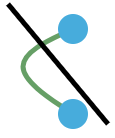} relations between both locutions and propositions. All colors correspond to the colors of the affected entities at the time of each interaction. 
The timeline reveals different annotation patterns that also became apparent in our expert user study: some users identify locutions and directly connect them with links wherever possible, while others focus on extracting all locutions first and create links in a second step.
The confirmation \faicon{check-circle} and rejection \faicon{times-circle} of locutions as well as annotation \inlinegraphics{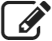} changes, including those to illocutionary connectors and argumentation schemes are displayed as well.

The timeline does not show any contextual information for the recorded annotation changes. According to the experts we consulted, this is not necessary as they are aware of the changes they performed during the last few minutes. However, they did express the wish to keep modifying the respective elements directly from the timeline, for example, to adjust the previous interaction based on a new insight. The interaction tracking serves as a data collection method for further improvements to \emph{VIANA}. While the current version uses ``locution interactions'' to improve suggestions, future versions of the system will use the data to provide undo and redo actions. 
Furthermore, the timeline is a first step towards providing analytic provenance.
\endgroup

\section{Interaction-Driven Annotation Suggestions}
\label{sec:learning_from_interaction}
To guide and support users in their analysis process, \emph{VIANA} highlights important keywords and recommends text fragments that should be annotated as locutions. While linguistic rules provide a good basis for initial suggestions, there is great potential for efficient human-machine collaboration. As there is generally not enough training data for good-quality classifiers or fully-automated annotation systems, \emph{VIANA} utilizes the knowledge encoded in a language model and refines it over time to train a measure of argument similarity from user interactions.

We employ a BERT~\cite{2018arXiv181004805D} model that was obtained by fine-tuning the ``BERT-Base'' checkpoint for evidence- and claim-detection. We gathered the training samples from several IBM Project Debater training datasets~\cite{Rinott2015ShowDetection, Shnarch2018WillMining, Aharoni2014ATopics} and chose the BERT base model because of its state-of-the-art performance. However, the recommendation component in \emph{VIANA} is not specific to BERT and could also be used with embeddings from any other language model.

The system first enumerates so-called ``elementary discourse units''~\cite{Hautli2017RhetoricalDialogs}--(sub)sentences delimited by punctuation and clausal connectors--as \emph{fragments}. Each fragment is represented as a tuple $(embedding, label, weight, points, state)$. Besides the embedding \textbf{$e$} the tuple contains a label \textbf{$l$}  $\in \{-1, 0,1\}$ that indicates whether a fragment should be an annotation ($l=1$), not be an annotation ($l=-1$) or is still undecided ($l=0$).
The weight \textbf{$w$} is used in similarity computations between two fragments. It is initialized to $w=1$ and updated through user interactions as described in \autoref{sec:suggestion_refinement}.
The points are initialized to $p=5$ and decay over time if users ignore suggested fragments but interact in their vicinity. The state $c$ of a fragment can be either \textsc{created}, \textsc{linguistics} or \textsc{confirmed}.

\emph{VIANA} only recommends locutions for annotation, not relations between them or their extracted propositions. While this is a field that we will explore in future work, discussions with annotation experts showed that they are more reserved with respect to proposed relations than proposed locutions. This skepticism stems from the fact that annotating relations is a significantly more complex task, and experts prefer to do a good job manually rather than having to correct imperfect suggestions. Therefore, we do not include proposed relations yet. 

\subsection{Initializing Suggestions}
\label{sec:initial_suggestions}
To avoid a cold-start of the recommendation system, we initialize it with the output of a linguistic discourse-unit annotation pipeline~\cite{Hautli2017RhetoricalDialogs} and refine it throughout the annotation process. 
The pipeline identifies discourse units that have a connection of type \emph{conclusion}, \emph{reason}, \emph{condition} or \emph{consequence} to the surrounding discourse units. We add all discourse units that contain a speech act of type \emph{agreement} or \emph{disagreement} and set the labels for the suggestions to 1 and their state to \textsc{linguistics}. All other discourse units are labeled 0 and remain in state \textsc{created}.

\begingroup
\setlength{\abovedisplayskip}{3pt}
\setlength{\belowdisplayskip}{0pt}
To suggest fragments to users, we compute the score $s$ for each fragment $f_i$ as the average of the weighted cosine similarity:  
\begin{align*}
s_i &= \frac{1}{j}\cdot\sum_{j \in C}cos(e_i, e_j)\cdot w_j \cdot l_j\cdot p_i
\end{align*}
where $C$ is the set of fragments f such that $C=\{f~|~f_c=\textsc{confirmed}~ ||\\f_c=\textsc{linguistics}\}$ and $cos(a,b)$ the cosine similarity between $a$ and $b$. Already (partly) decayed points $p_i$ and a higher similarity with fragments $f_j$ with a negative label $l_j$ lead to a lower score, and hence a lower probability of $f_i$ being suggested as an annotation. 

We chose cosine similarity over the dotproduct, and normalization factor $\frac{1}{j}$ over $\left(\sum_{j\in C}w_j\right)^{-1}$ as this combination led to the best separation between confirmed and rejected suggestions in our tests. 
\endgroup

After sorting all fragments with $c=\textsc{created}$ according to their score we return the $n$ suggestions with the highest score. 
Any fragments with $c=\textsc{linguistics}$ are shown as suggestions (if they have not been manually deleted), independent of their score.
While the correct number of $n$ depends on the datasets, our study participants noted they wanted rather more than fewer suggestions.

\subsection{Promotion and Decay of Suggestions}
\label{sec:suggestion_refinement}

To refine the suggestions over time, we update the weights and labels of fragments through user interaction. When users confirm or disconfirm (i.e., mark as a draft) a locution, we triple the weight of the associated fragment or divide it by three, respectively. Deleting a locution leads to the weight of the associated fragment being doubled, while the label is changed from $1$ to $-1$. Manually added locutions are initialized with a weight of $4$. 
These weights and updates ensure that items that have been interacted with take a more important role in the similarity calculation. Compared to confirmed locutions we keep the weights for deleted locutions lower to ensure that the system keeps proposing fragments that should be annotated, rather than those that should not be. The concrete values of the weights were identified through initial experimentation. Their general distribution (lowest values for negative feedback, highest values for manual intervention) follows our previous work on model optimization through progressive learning~\cite{El-Assady2018}.

To avoid cluttering the screen with suggested annotations, we introduce a \emph{viewport-dependent suggestion decay} function. While there are simple ways to learn from direct user-interaction as described above, the number of interactions in a system is limited.  \emph{VIANA}, thus, also learns from the items that users do not interact with. Recall the points $p$ associated to every fragment. Whenever users interact with an item, suggestions that are close on screen but are not interacted with, lose some points. This process captures that the suggestion was not relevant to the users, and they rather performed a different action. We calculate the point loss of fragment $f_i$ after an interaction with fragment $f_j$ as $pl_i = \log_{32} D - \log_{32} d(f_i, f_j)$ where $D$ is the maximum (absolute) decay distance and $d(a,b)$ the distance in the text as number of words between fragments $a$ and $b$. 
The maximum decay distance $D$ is dependent on the amount of text visible on screen and ensures that only those suggestions that are visible to users and likely to be ``interaction alternatives'' lose points. \emph{VIANA} currently sets $D$ to 200. As a result, each fragment can lose, in theory, at most 1.5 points after each interaction. In practice, the maximum loss is closer to 1 due to the typical length of a locution.
We update the points of visible suggestions after each interaction, with the exception of confirming suggestions.

Once a fragment lost all points, we set its weight to 2, its state to \textsc{confirmed}, and its label to $-1$, mimicking a user manually deleting the suggestion. Fragments that lose all points are taken into consideration as negative samples when generating new suggestions.
This novel approach for viewport-dependent suggestion decay differs from previous work on suggestion decay in recommender systems that is typically based on temporal evolution or ignores on-screen context. Temporal decay is not suitable for argumentation annotation as there are no changes to annotation guidelines during an individual annotation. Our approach penalizes individual items in addition to updating the content-based similarity function.
It incorporates the on-screen distance between ignored suggestions and interactions to inform the speed of decay. This takes into account that users are likely to be much more aware of the content in the direct vicinity of their interaction, especially in text-based systems.

\section{Evaluation}
To validate the effectiveness of our approach, we conducted an expert user study with five participants over a period of three months. We present its results after introducing two independent use-cases that demonstrate the usefulness and practical applicability of \emph{VIANA}. 

Expert E2 (who will be introduced in the following section) validated the choice to forgo a quantitative evaluation and stated that inter-annotator agreement studies for argumentation often fail because of the multi-stage process of annotation. Once annotators disagree in the identified locution structure, their propositions, relations or argumentation schemes are automatically not in accordance. 

Furthermore, the time needed for annotation is not necessarily a useful metric as experts already spend many hours annotating and prefer exact over fast results. Much time is spent on reasoning about the underlying argument structure during annotation. Consequently, computing speed-up factors between two subsequent annotation runs in different annotation systems is meaningless. As the argumentative structure will have been identified during the first annotation, the second one will always be easier. Hence, we present qualitative feedback from five study participants.  In particular, we highlight their feedback with respect to the design and usability of the system, as well as the usefulness and quality of the suggestions.

\subsection{Use Cases}
Before providing qualitative feedback from our expert user study, we present two use-cases that highlight the general usability of \emph{VIANA}. 

\subsubsection{Argumention Annotation}
When loading \emph{VIANA} to annotate the second presidential debate between Obama and Romney in 2012, the user first sees the close reading view. She begins to read until she finds some utterances on taxes that she decides to annotate. As soon as she begins annotating ([T2]), she is presented with automated suggestions like those shown in \autoref{fig:slow_analytics_view}. As she finds the annotation suggestion for ``The top 5 percent will continue to pay 60 percent'' helpful and decides to accept it, she immediately sees updated suggestions. After annotating some locutions, she begins to introduce relations between them ([T3]) and later switches to the \emph{Argument Reconstruction View}. There, she reconstructs propositions ([T4]) to fix grammatical errors, remove trailing punctuation and turn questions into statements. Next, she investigates the argumentative structure of the annotated segment. The yellow rephrase relation highlights Romney's plan to lower taxes. Using her scroll wheel to progress to the \emph{Argument Exploration Map}, the user finds that the debate also seems to discuss college education in the context of taxes and the economy ([T5]). She decides to investigate this topic more closely and figure out how additional arguments relate to those already identified. Switching back to the \emph{Locution Identification View} she soon finds an exchange between Obama and an audience member on tax reduction through education credits that she annotates next. 

\subsubsection{Argumentation Behaviour}
\begin{figure}
    \centering
    \includegraphics[width=\linewidth]{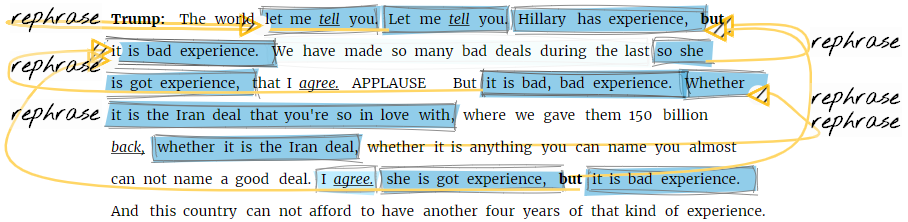}
    \caption{One utterance by D. Trump from the first Presidential Debate between him and H. Clinton in 2016. His constant rephrasing is made obvious by the sheer amount of yellow edges on screen. }
    \label{fig:trump_rephrasing}
    \vspace{-\baselineskip}
\end{figure}
Moving on to the first presidential debate between Hillary Clinton and Donald Trump, we can use the annotation of argumentation to observe different argument behavior strategies. 
\autoref{fig:trump_rephrasing} shows a single utterance by Donald Trump that a user annotated in \emph{VIANA}. Trump is attacking Hillary Clinton by claiming that her experience as Secretary of State is ``\textit{bad experience.}'' He then keeps repeating and rephrasing this claim through the entire utterance. While he occasionally mentions other facts or makes new claims, these are not connected to other propositions with inference or conflict relations. Instead, he repeatedly restates his opinion on her qualifications. 

In such a scenario, displaying the annotation directly in the text view yields valuable insights. It is easy to see that the utterance is intra-connected with rephrases, while still containing locutions without any relations. The text view clearly communicates this finding as the close packing of the locutions shows the immediate context of each rephrase. A graph view containing only locutions or propositions would likely fail to transmit the same image due to the lack of textual context. As a consequence, it would not become immediately obvious that all annotations originated from a single utterance, or whether long breaks separated the annotated fragments. 

Similar phenomena can be observed when discussion participants make unsubstantiated and unrelated claims or contradict themselves within the context of a single utterance or a short period of time. 

\subsection{Expert User Study}
In addition to the use-cases presented above, we evaluate \emph{VIANA} in a two-stage Expert User Study carried out in pair analytics sessions~\cite{Kaastra2014FieldAnalytics}. In each session, one domain expert and one visual analytics expert (one of the authors) were present. The two stages of the study were performed three months apart. The system evaluated in the second stage incorporates expert feedback from the first study.

\subsubsection{Methodology}
We divided the 90-minute long study sessions into three parts. In the first 20 minutes, we presented the system to the expert and explained the functionality. We also elicited initial feedback on the design-choices and usefulness of \emph{VIANA} through a semi-structured interview. In the following 40 minutes, we gave the expert control over the system interface and let them explore the dataset. We supported the expert with clarifications on the functionality of the user interface and the controls whenever they had questions. Occasionally we also proposed to progress to a different view to ensure that the expert got a holistic impression of the system. Each expert was encouraged to think aloud and explain the rationale for their actions. 
After the exploration period, we transitioned into another semi-structured interview of 30 minutes. Here we asked the expert for detailed feedback based on their experience with the system to receive a general assessment. We focussed on the design, the usefulness of the tool to the expert, the quality of suggested annotations, and potential missing features. 
We recorded both audio and video from the screen during all study sessions.

\textbf{Participants -- }
\textbf{E1} holds a Ph.D. in computational linguistics and currently works as a postdoctoral researcher. As argumentation has a crucial role in her research, she currently spends multiple hours a day annotating argumentation data. She also teaches courses about argumentation and the underlying theory. \textbf{E2} just completed his Ph.D. at the intersection of computer science and computational linguistics, working on argumentation and ethos mining. 
He estimates to have worked on manual argumentation annotation for over one full year in the last 3.5 years of his Ph.D.
During the design phase, E2 provided his domain knowledge about less common IAT annotations that he felt should be possible in \emph{VIANA}. As a consequence, he had seen, but never used, the system before the study. He did not contribute to the visual design, the interaction design, or any of the analysis layers in particular.

Participants \textbf{S1}, \textbf{S2}, and \textbf{S3} are masters students in Speech and Language Processing, have received special training in argumentation annotation over six months and work as student assistants in argumentation annotation now. 
E1 and E2 participated in the first, and S1, S2, and S3 in the second phase of the study. None of the experts are authors of this paper.

\textbf{Dataset -- }
As datasets, we chose transcripts of presidential debates as all experts had experience with annotating political debates. In the first study phase, we used the second 2012 debate between Barack Obama and Mitt Romney because E2 had previously annotated all three of the more recent debates between Trump and Clinton. In the second phase of the study, we used the first 2016 debate between Trump and Clinton. None of the experts had annotated the respective data used in the study before. All participants were presented with 40 utterances from the end of the debates. This section of the text has been selected to skip the non-argumentative introduction phase of the debate and fit the text-length to a typical annotation session that our experts are used to. 

\subsubsection{Results and Feedback}

In this section, we report the feedback from the three phases of our study sessions. We summarize the comments of the participants, providing a selection of the most insightful feedback.

\textbf{Initial Feedback -- }
Both E1 and E2 highlighted the importance of manually annotated argumentation for their research. None of the five experts had previously worked with visual analytics systems for argumentation annotation. However, they were excited about the idea of working with the proposed locutions, with E2 stating ``\emph{I think this would speed up the whole process [of annotating] quite significantly}.'' S1 agreed and articulated ``\emph{I wish I just had to check if the locutions were already extracted correctly.}'' E1 highlighted the importance of human-in-the-loop analysis. She liked the idea of proposed locutions ``\emph{as long as I can change them}.'' This response mirrors the generally reserved attitude of the experts towards fully automated approaches. At the same time, all experts expressed that they were aware that suggested annotations might bias the result towards the suggestion, especially in hard-to-decide situations. 

While she thought the extracted tasks were relevant and captured her actual work, E1 was skeptical of the individual analysis layers we introduce for each task. As her recent work has often focussed on illocutionary connections, she stated to be ``\emph{suspicious about the illocutionary structure}.'' S2 had similar skepticisms at first. However, she changed her opinion after being introduced to the system and mentioned she thought ``\emph{it is more manageable.}'' S3 agreed that it is ``\emph{good to see the different layers. Sometimes it makes things easier to understand.}'' One of E2's biggest complaints about the system he currently uses is that ``\emph{if you do not constantly move nodes around to make it look good you lose track of what is going on}'' and that ``\emph{significant time is spent making the graph readable}.'' Consequently, he liked the idea of automated layouts and separated layers introducing a more rigid visual structure to the graph. Later system use revealed two annotation patterns. While some users directly interconnect extracted locutions, others first extract some locutions, before transitioning through the layers to link and reconstruct them. They then transitioned back to the text view, effectively annotating data in ``mini-batches''. 

\textbf{Design and Usability -- }
When presented with the system for the first time E1 remarked that the sketchy-rendering of nodes and relations might be confusing and could suggest ``\emph{that you are not actually sure what you are doing.}'' She argued that she was performing a ``\emph{precise analysis}'' that did not warrant a sketchy representation. In her expectation, only proposed locutions should have been sketchy. This sketchiness should then have been removed once users confirmed the entities or never be introduced for those created manually. E2 was more fond of our design choice and described the visual design as ``\emph{modern.}'' He also felt that the sketchy rendering made him more inclined to keep editing the annotation as opposed to existing systems where he feels like no changes can be made anymore. As a reaction to these opinions, we made the sketchiness configurable for the second phase of the study. Here, S3 chose the sketchy version because it felt ``\emph{just like working on a piece of paper}'', while S2 dismissed the sketchy option saying ``\emph{it does not feel as official.}''

Both S1 and S2 were concerned with the superposition of the extracted locutions and the text, unanimously calling it ``\emph{very dense.}'' They later stated that they preferred the offered graph views to create new links between arguments. After using the system, S2 felt that the design was ``\emph{not cramped}'' and that the ``\emph{highlighted parts seem clearly separated.}'' Additionally, she now thought the superposition made it easier to add, delete, or refine locutions. In the current version of the system, users have to remove irrelevant suggestions manually or wait for them to decay over time. S3 wanted the system to automatically remove suggested annotations as soon as she manually annotated an overlapping locution, a request that we plan to include in future work. 

\textbf{General Assessment -- }
The experts unanimously praised the system for its proposed locutions and the potential they bring for reducing the overall time needed for annotation. They did also have ideas for small improvements, like resizing an already identified locution to add or remove individual words from the beginning or end. S1 noted that having suggestions ``\emph{makes things a lot easier.}'' However, she also noted that a careful trade-off had to be made between showing too many and too little suggestions and matching the users' expected density. In the subsequent study, S2 reached three consecutive sentences without suggestions and exclaimed ``\emph{it makes me wonder whether I am over-annotating.}'' As she continued annotating suggestions appeared in the previously empty area, leading her to conclude that the system was learning from her interaction. 

After her annotation session, S2 began to reason about the quality of suggestions. She liked ``\emph{that \emph{[VIANA]} has two `tracks' of suggestions}'', referring to the different colors assigned to suggestions from the linguistic pipeline and the learned user interactions and said she was ``\emph{trying to figure out which ones [she] trust[s] more.}'' She concluded that she found the linguistic suggestions more reliable in the beginning, but would become more reliant on those learned from her interactions over time. E1, S1 and S2 expressed that they did not feel that the suggestions had biased the annotation result, with S1 saying ``\emph{you still need to think about the suggestions.}''

While none of the experts found the \emph{Argument Exploration} view essential during the annotation of a section of text, E1 stated she found it useful to communicate the results to colleagues. E2 liked it in particular as a means to introduce long-distance relations between nodes that would be far apart in the other views. He also imagined using it whenever coming back to the annotation after a break to get back into the context quickly. S3 found the map ``really useful'' when annotating longer texts and envisioned that you could ``use it to check on yourself'' and observe your progress.

Summarizing her experience with the system, computational linguist expert E1 said ``\emph{This is really nice, and it will help a lot.}'' One important factor for her was the fact the \emph{VIANA} enables her to load and annotate larger amounts of data at once: ``\emph{I definitely like that you can put in a lot of text.}'' This saves time that is otherwise spent on combining the results of the annotation of multiple chunks. However, she was not in favor of splitting the locutions and propositions onto two different analysis layers, preferring a side-by-side view instead of having to switch layers during the annotation process, as this would be ``\emph{distracting}.'' She stated ``\emph{we want locutions and propositions next to each other}'' and suggested introducing a fifth layer in the middle of our current set of views. We have since added such a combined view after the first phase of the study. E1 also expressed that she would still use the other views before and after the annotation process, naming the note-taking interface (``\emph{it would be really great to have [VIANA], also because of the comment functionality}'') and the topic map to communicate the final results.

After the study had ended, E2 remained seated in front of the system and kept transitioning between the layers, saying ``\emph{this is so cool!}'', validating our approach of bringing visual analytics to argumentation annotation. While the expert users have strong mental models reinforced through the long-time use of existing systems, they are open to new developments. Despite the learning period of new systems, it is promising that users found it engaging in the little time they had with the system. 

\subsection{Discussion and Lessons Learned}
The expert user study highlighted the demand for a flexible tool that is capable of adapting to the users' needs and expectations.  While they are interested in new systems and eager to try them out, they do have very specific layouts and functionalities in mind. It is only through personalization of the interface and exactly understanding their work processes that we can provide them with an efficient system~\cite{Hinrichs2017RiskVisualization}. The current version of \emph{VIANA} makes the \textit{sketchiness} of the tool configurable, giving users the option to keep it active at various degrees of intensity or disable it outright. As the annotation process is subjective and highly time-consuming, we need to ensure that we cater to the different mental models of users to create a good user experience for them. 

Users already liked the relatively high degree of automation that we provide. Despite being generally reserved concerning fully-automated argumentation mining tools, the experts were more at ease once they knew that they would still be able to manually ``overwrite'' the system later. The general wish for more automation became apparent when S1 asked ``\emph{do I need to do this manually?}'' when changing an illocutionary connector. She was already comfortable with the interaction model of suggested locutions and would have preferred to confirm a suggested change here as well. Summarizing all study sessions, the list of requested automation steps includes the resolution of reported speech and pronouns, the lowercasing of propositions, selection of argumentation schemes, or ``hiding'' non-argumentative  areas of text.

As the pre-usage interviews revealed, all experts were aware of potential bias introduced by suggestions. None of them did, however, feel that their final annotation was biased. Consequently, future research is needed to determine whether, and to what degree, users are exposed to various kinds of bias.
The study showed not only the effectiveness of our learning approach but also trust-building processes facilitated by simple interaction principles. Before using the system, most users were unsure about the quality of suggestions. After performing a few interactions and realizing that they had full control, they started to build trust and looked forward to the next suggestions, showcasing a successful human-machine-collaboration.

\textbf{Limitations -- } 
The design of any system is often a trade-off between expressiveness, complexity, and ease of use. Using layers and semantic transitions, we aim to reduce the complexity of the system and make it intuitive to use. Our evaluation shows that some users have existing workflows that are not well-suited to such a layered approach, while others preferred the reduced complexity. As \emph{VIANA} provides various layers and combinations of views, users can customize their workspace to their tasks and needs.
The current implementation of \emph{VIANA} is tailored towards texts of up to 10,000 words and thus suitable for typical text lengths in argumentation annotation, according to experts. The argument exploration view becomes crowded for longer texts and would require additional navigation interactions to accommodate the space requirements. Furthermore, annotation is still a manual process and does consequently not scale to very large amounts of data.
With increasing text length and, more importantly, increasing annotation density, views like \autoref{fig:trump_rephrasing} become more typical.  An even denser annotation is theoretically possible but is not likely on real data in an expert-user system.
In future work, we will investigate grouping of arguments to enable focusing on particular areas of the data. Showing and hiding such groups can enable scaling to larger datasets. Further, we plan to utilize the design element of sketchiness to encode (non-continuous) information. 

\section{Conclusion}
We have presented \emph{VIANA}, a web-based, integrated approach that enables both the annotation of argumentation, as well as the exploration of the results. It provides stacked, task-specific interface layers that we connect with smooth semantic transitions.
\emph{VIANA} automatically suggests fragments of text for annotation and lays the foundation for an extensible platform that will be developed towards semi-automated argumentation annotation and argument mining. The suggestions are refined over time by learning from both the presence and absence of user interaction, introducing a novel approach to suggestion decay.
The web-based architecture of \emph{VIANA} makes it easy to distribute it to new users and opens possibilities for further extensions towards a remote-collaborative tool with which multiple users can annotate individual segments of text at the same time. This is especially interesting in fast-paced environments like the live annotation of radio- or TV-shows. 
\emph{VIANA} will be made available as part of \href{http://lingvis.io}{lingvis.io}~\cite{lingvisio} under  \href{http://viana.lingvis.io}{http://viana.lingvis.io}.

\acknowledgments{
This work has been funded by the DFG with Grants 350399414 and 376714276 (VALIDA/SPP-1999 RATIO).
}

\bibliographystyle{abbrv-doi-hyperref}
\bibliography{template}

\end{document}